\documentclass[referee,iicol]{sn-jnl}

\usepackage{graphicx}%
\usepackage{multirow}%
\usepackage{amsmath,amssymb,amsfonts}%
\usepackage{mathrsfs}%
\usepackage{xcolor}%
\usepackage{textcomp}%
\usepackage{manyfoot}%
\usepackage{booktabs}%
\usepackage{algorithm}%
\usepackage{algorithmicx}%
\usepackage{algpseudocode}%
\usepackage{listings}%
\usepackage{nth}%

\begin{document}

\title[Article Title]{Offline Detection of Misspelled Handwritten Words by Convolving Recognition Model Features with Text Labels}

\author{\fnm{Andrey} \sur{Totev}}

\author{\fnm{Tomas} \sur{Ward}}

\abstract{Offline handwriting recognition (HWR) has improved significantly with the advent of deep learning architectures in recent years. Nevertheless, it remains a challenging problem and practical applications often rely on post-processing techniques for restricting the predicted words via lexicons or language models. Despite their enhanced performance, such systems are less usable in contexts where out-of-vocabulary words are anticipated, e.g. for detecting misspelled words in school assessments. To that end, we introduce the task of comparing a handwriting image to text. To solve the problem, we propose an unrestricted binary classifier, consisting of a HWR feature extractor and a multimodal classification head which convolves the feature extractor output with the vector representation of the input text. Our model's classification head is trained entirely on synthetic data created using a state-of-the-art generative adversarial network. We demonstrate that, while maintaining high recall, the classifier can be calibrated to achieve an average precision increase of 19.5\% compared to addressing the task by directly using state-of-the-art HWR models. Such massive performance gains can lead to significant productivity increases in applications utilizing human-in-the-loop automation.}


\maketitle

\section{Introduction}\label{sec1}

Handwriting recognition (HWR) models extract text from data representing a person’s handwriting. The two main variants of the task are offline HWR which operates on handwriting images and online HWR which receives temporal sequences of coordinate points denoting the pen’s trajectory and state.

HWR models are typically evaluated using character error rate (CER) and word error rate (WER) which are obtained from the Levenshtein distance between the predicted text and the ground truth \cite{b1} using the following equations:

\begin{equation}
CER = \frac{S_c+I_c+D_c}{L_c}\label{eq:CER}
\end{equation}

\begin{equation}
WER = \frac{S_w+I_w+D_w}{L_w}\label{eq:WER}
\end{equation}
where $S_u$, $I_u$, $D_u$ ($u\in\{c, w\}$) represent the number of substituted, inserted, and deleted character or word units in a ground truth text with a length of $L_u$ units.
Unconstrained state-of-the-art models, such as the ones based on CNN-RNN architectures, achieve a CER of 5-10\% and a significantly higher WER of 15-20\% \cite{b2,b3}. This is understandable because 1) a single character is much more likely to be predicted correctly than a whole word and 2) the relative contribution of a wrongly predicted word to the error rate is greater due to the lower number of words than characters in any test set.

Restricting the HWR predictions via lexicons or language models leads to significant improvements in the accuracy of the models, reducing CER to below 5\% and WER to 5-10\% \cite{b2,b3}. One side effect of such approaches, however, is the implicit ``correction'' of the recognized text. This behaviour is suitable for offline models which normally perform data extraction tasks where the information may be indexed for searching and the intended meaning is more important than capturing involuntary mistakes. By contrast, applications that aim to recognize the handwriting precisely and without modifications, are typically used for spell checking and utilize online (rather than offline) models to help users identify their mistakes while writing.

\begin{figure}[htbp]
\includegraphics[width=0.48\textwidth]{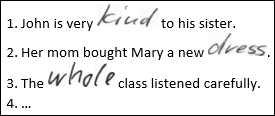}
\caption{Dictation exercise}\label{dictation}
\end{figure}

\begin{figure}[htbp]
\includegraphics[width=0.48\textwidth]{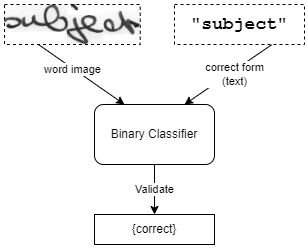}
\caption{Our proposal - a one-step process using a binary classifier}\label{classifierValid}
\end{figure}

\begin{figure}[htbp]
\includegraphics[width=0.48\textwidth]{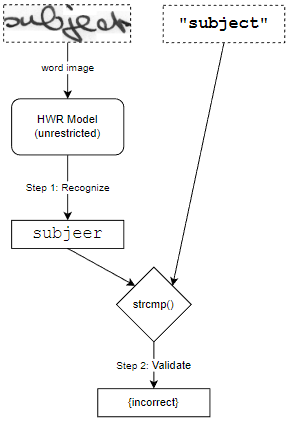}
\caption{Conventional approach - a two-step process using a generic HWR model}\label{hwrValid}
\end{figure}

Despite the extensive research in areas such as recognition and spell checking of handwritten text, to the best of our knowledge, none of the literature investigates the problem of validating handwritten text against its known correct form (Figure~\ref{dictation}, Figure~\ref{classifierValid}). This task is quite common in a school context (e.g. to assess student answers by looking them up in an answer book) and is uniquely positioned in the HWR space: to remain sensitive to mistakes, a solution cannot rely on language constraints, however it still needs to be language-aware given the textual input received alongside the graphical image input. Due to the novelty of our task and the lack of existing solutions, we evaluate it against a conventional approach where the handwriting image is input to a state-of-the-art HWR model and the predicted text is compared against the correct text (Figure~\ref{hwrValid}).

To address the handwritten text validation problem, we propose the following contributions:
\begin{itemize}
\item A one-step process for detection of misspelled words using a binary classifier.
\item A novel neural architecture for the binary classifier utilizing a HWR model as a feature extractor and a multimodal classification head which convolves the feature extractor output with the vector representation of the input text.
\item A training process relying entirely on synthetic data created using a generative adversarial network (GAN) pre-trained on the same dataset as the feature extractor.
\item We demonstrate that, while maintaining high recall, our model achieves an average precision improvement of 19.5\% against a conventional method involving a state-of-the-art HWR model.
\end{itemize}

In the remaining parts of this work, we first introduce the main model architectures for offline HWR and the most popular handwriting datasets. We also review common approaches to generating realistic handwritten text and their applications. Next, we present a binary classifier for detecting misspelled words together with a training method relying entirely on synthetic data generated using a GAN architecture. Finally, we evaluate our system in a practical context and compare it against the two-step baseline approach from Figure~\ref{hwrValid}.

\section{Related Work}\label{sec2}

HWR research is predominantly focused on developing improved artificial neural network (ANN) architectures for addressing the recognition task. Models such as \cite{b2,b3} achieve state-of-the-art performance via a sequence of convolutional and recurrent layers with a connectionist temporal classification (CTC) layer for decoding the output sequence. As \cite{b3} demonstrated, recognition models show superior performance when constraining the prediction using a lexicon at the expense of recognizing out-of-vocabulary (OOV) words. This can be partially mitigated by employing language models of multigrams \cite{b4} and character n-grams \cite{b7}. (By contrast, our solution needs to be free of language constraints in order to remain sensitive to mistakes which cannot be captured by a language model.)

\subsection{Main Architectures}\label{subsec2}

\subsection{CNN-RNN}\label{subsubsec2}

The state of the art is dominated by architectures combining CNN with one-dimensional LSTM layers and using a CTC output layer \cite{b2,b3,b4}. Compared to earlier multi-dimensional LSTM architectures proposed by \cite{b15,b17,b19}, these CNN-1DLSTM models achieve increased computational efficiency and significantly improved accuracy when using data augmentation \cite{b10}.

The decoding part of CNN-RNN architectures is commonly enhanced by utilizing a lexicon which constrains the output to known or learnt sequences. \cite{b7} increases the robustness by predicting n-gram hit lists before calculating the total probability for each matched lexicon entry. \cite{b14} proposes an ultra-wide-lexicon method devising a ``cohort'' of LSTM verifiers - each one selected at the end of a training epoch. \cite{b18} uses a similar approach of ``cascaded'' complementary LSTM classifiers with identical architecture but trained using different random weight initialisation.

A variation of CNN-RNN class of models follows the Seq2Seq approach, introduced by \cite{b39}. \cite{b16} replaces the CTC decoder with an LSTM decoder with attention whereas \cite{b20} augments the CNN component with a self-attention block.

\subsection{Non-recurrent}\label{subsubsec3}

Several authors propose recurrent-free approaches with state-of-the-art performance which allow for greater training speeds via parallelisation. \cite{b12} employs the YOLOv3 object detection model in a lexicon-free solution which trains using significantly less data. \cite{b11,b13} also devise fully convolutional architectures achieving state-of-the-art results. More recently, \cite{b4_2,b9} and other authors propose Transformer-based architectures with an optional language model.

\subsection{Datasets}\label{subsec3}

The importance of IAM \cite{b37} and RIMES \cite{b38} datasets cannot be overstated. Nearly every publication in the space of offline HWR uses them for training and evaluation purposes. The images in IAM and RIMES are collected using flatbed scanners which makes them less practical in a modern context where most content is digitised using mobile devices. GNHK \cite{b4_1} aims to address this gap by providing a collection of handwriting images taken only with smartphones. Additionally, Transformer-based models are reported to massively overfit the IAM dataset \cite{b9} posing a need for larger benchmarking datasets in the era of large neural network models.

\begin{figure}[htbp]
\includegraphics[width=0.48\textwidth]{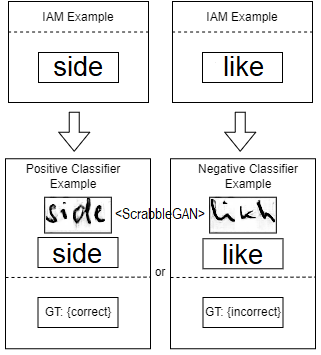}
\caption{Deriving positive and negative examples from a HWR benchmark dataset by injecting noise and generating the handwriting image using ScrabbleGAN}\label{exampleForming}
\end{figure}

Another important limitation of the available handwriting databases, is the complete lack of annotated misspelled examples. Since the models in this work are unconstrained by lexicons or language models, we leverage this by creating the misspelled examples: the original ground truth label is altered and the corresponding handwriting image is generated as depicted in Figure~\ref{exampleForming}.

\subsection{Handwritten Text Generation}\label{subsec4}

One of the main applications of handwritten text generation is to provide synthetic data for training handwriting recognizer models without aiming at mimicking an individual’s handwriting style \cite{b54_1,b54_2,b54_4}. Other models, which can also be used for training HWR models, are conditioned on style \cite{b54_3,b54_5,b54_6,b54_7}. The majority of the work in synthetic handwriting is based on generative adversarial networks (GANs). In this work, we train our classifier on examples generated using ScrabbleGAN \cite{b54_2} - a synthetic handwriting generator trained on the IAM dataset which allows the manipulation of the resulting style using parameters.

\section{Proposed System}\label{sec3}

\begin{figure}[htbp]
\includegraphics[width=0.48\textwidth]{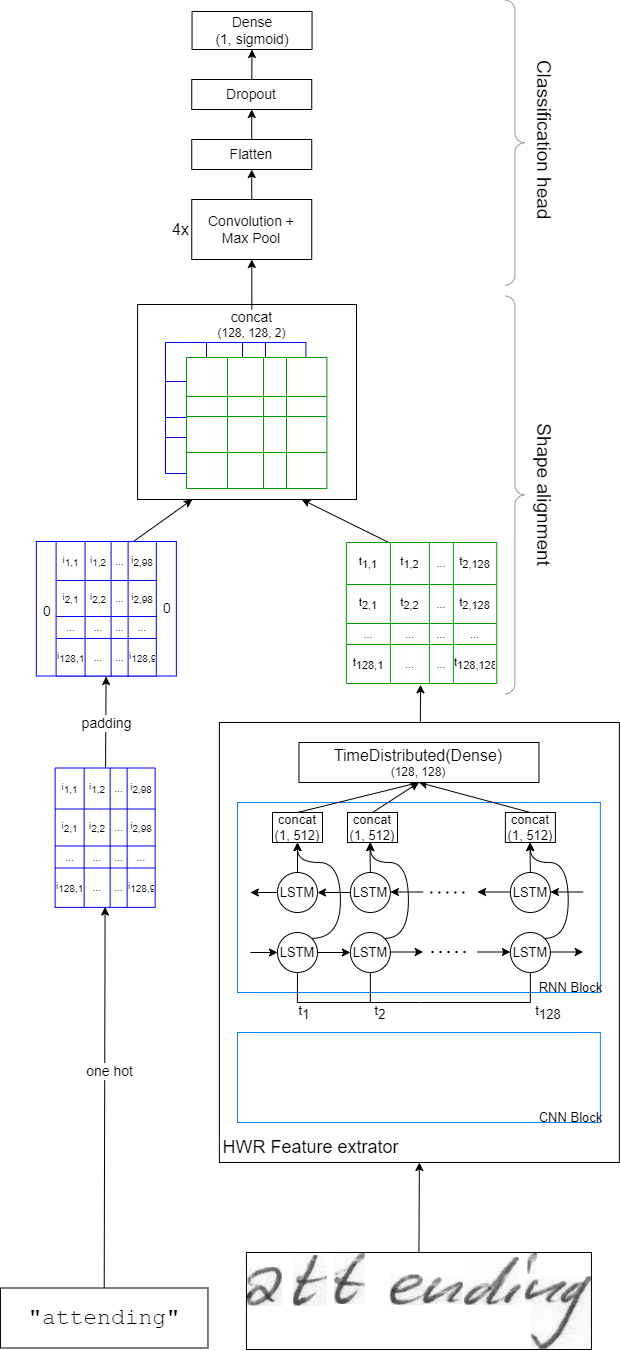}
\caption{Architecture for offline detection of misspelled handwritten words using feature extractor based on \cite{b2} and a custom convolutional classification head}\label{systemOverview}
\end{figure}

An outline of the system is provided in Figure~\ref{systemOverview}. In this section we describe the architecture, the training process, and relevant implementation details.

\subsection{Architecture}\label{subsec5}

\subsection{HWR feature extractor}\label{subsubsec4}

We pre-train the CNN-RNN model described in \cite{b2} on the IAM dataset. It consists of a series of convolutional blocks followed by a bidirectional LSTM block which feeds its output sequences into a fully-connected layer classifying the characters in the output sequence. For the purposes of our binary classifier, we discard the top fully-connected layer and use the lower layers for feature extraction.

\subsection{Aligning the shapes}\label{subsubsec5}

The RNN block consists of 128 time steps each outputting a 512-long sequence. We shrink the output sequences to 128 units by using a time-distributed dense layer and stack them into a 128x128x1 tensor. Similarly, we stack the one-hot encoded input text into a 128x98 tensor and apply symmetric zero-padding to shape it into a 128x128x1 tensor. The two tensors are concatenated into a 2-channel ``image'' which is fed to the classification head where the HWR features are convolved with the correct text vector in order to detect patterns indicating similarity between the two representations.

\subsection{Convolutional classification head}\label{subsubsec6}
The classification head contains only 135041 trainable parameters. It consists of four 3x3 convolutional blocks with same padding and 2x2 max pooling. The output of the final block is flattened, a dropout rate of 0.1 is applied, and the result is fed into a perceptron with sigmoid activation. The main motivation behind using a convolution here is to leverage the spatial correlation that intuitively exists between the HWR feature matrix and the one-hot matrix. (The HWR models are trained to predict the correct characters and their order, hence the correlation.) Additionally, removing the top HWR layers and using the output of the lower layers as features, allows our model to learn broader and richer relationships with the text vectors due to the increased receptive field of the convolution.

\subsection{Training dataset}\label{subsec6}

In this work we derive our training examples by sampling 100000 words from the IAM test set and generating a randomly styled handwriting image of each word using ScrabbleGAN. Binary classification models typically benefit from balanced training sets, therefore in 50\% of the training examples we inject noise (similarly to the ``SpellMess'' method \cite{b55}) by altering a randomly selected character in the label with a random lowercase alphabetic character before generating the handwriting image. As denoted in Figure~\ref{exampleForming}, the handwriting image matches the text in the examples labelled as "correct" and doesn't match the text in the "incorrect" examples. The early stopping is controlled via a balanced validation set of 15000 examples derived from the IAM validation set using the same scheme. The noise injection in the negative training examples is based on the procedure from Algorithm~\ref{mistakeGen} \cite{b50}. The main advantages of the method are its simplicity, computational efficiency, and the ability to improve the model sensitivity to even small mistakes without relying on any language features.

\begin{algorithm}
\caption{Generation of mistaken words}\label{mistakeGen}
\begin{algorithmic}[1]
\renewcommand{\algorithmicrequire}{\textbf{Input:}}
\renewcommand{\algorithmicensure}{\textbf{Output:}}
\Require word
\Ensure mistaken word
\State pos := random\_int(0, word.length())
\State new\_chr :=  random\_lowercase()
\State \Return word.replace(pos, new\_chr)
\end{algorithmic}
\end{algorithm}

\subsection{Training}\label{subsec7}

The HWR model is pre-trainined on the IAM ``line'' examples in accordance with the procedures outlined in \cite{b2} achieving best performance in the \nth{64} epoch. The feature extractor is then frozen, its character classification layer is replaced with the convolutional classification head. The new model is trained on the generated dataset using RMSprop optimizer and binary cross-entropy loss for 108 epochs progressively decreasing the learning rate from 0.001 to 0.00004.

\subsection{Implementation details}\label{subsec8}

Our implementation is based on the Tensorflow 2.0 HWR training and evaluation pipeline developed by \cite{b40}.

\section{Experimental Evaluation}\label{sec4}

\subsection{Dataset}\label{subsec9}

Similarly to how the training data is produced, the test sets are generated using ScrabbleGAN by sampling words from IAM's test partition. It is worth noting that the baseline HWR recognition model achieves comparable rates of correctly recognized words with both the original IAM test set (63\%) and a generated dataset sampling from the same words (62\%).

\subsection{Baseline approach}\label{subsec10}

The baseline approach consists of the following steps:
\begin{enumerate}
\item Apply state-of-the-art HWR model to a word image from the test dataset
\item Compare the predicted text with the ground truth label
\item The example is classified as a spelling mistake (i.e. positive) if and only if the predicted word differs from the ground truth label
\end{enumerate}

\subsection{Evaluation method}\label{subsec11}

For the purpose of our evaluation, we devise a dictation assessment scenario, similar to the exercise from Figure~\ref{dictation}, where a teacher reads out loud to the students who are tasked with filling the gaps in a test sheet.

Assuming normal distribution of the mistakes count across all students, we introduce two assessment levels - Moderate and Difficult - with mean mistake-to-correct word ratio of 1:5 and 1:2 respectively (Table~\ref{scenarios}). We also specify a minimal recall requirement of 99\%, indicating that only 0.2 counts of mistaken words remain undetected in an average 20-word assessment.

\begin{table}[htbp]
\caption{Specification of dictation assessment scenarios}\label{scenarios}%
\begin{tabular}{@{}llll@{}}
\toprule
                   & Mean         & Min. recall& Undetected\\
                   & mistakes     &            & mistakes\footnotemark[1]\\
\midrule
Moderate scenario  &       16.67\%&        99\%& 0.03 words\\
Difficult scenario &       33.33\%&        99\%& 0.07 words\\
\botrule
\end{tabular}
\footnotetext[1]{In a 20-word assessment.}
\end{table}

We further introduce the concept of mistake severity where each severity level (1-3) corresponds to the number of recursive invocations of Algorithm~\ref{mistakeGen} for generating the mistaken variant of a word. (The severity level is not necessarily equal to the number of mistaken symbols in the word as each invocation makes a random choice about the position in the string which should be altered.) Given the two assessment scenarios and three severity levels, we assemble six separate test sets which share the same positive examples but differ in the count and severity of the negative ones. We test our model on all datasets, but for simplicity group the results by assessment scenario (difficult, moderate) and average them across all severity levels.

Based on the scheme from Figure~\ref{exampleForming}, we form six distinct test sets. That is, for each test set we independently sample $T=15000$ words (with replacement) from the IAM test set and introduce mistakes at the required rate and severity level using Algorithm~\ref{mistakeGen}. These synthetic test sets compensate for the lack of annotated misspellings in the public handwriting databases and are suited to unconstrained solutions like ours, where the injected noise cannot ``confuse'' any language-based components

Since our baseline approach relies on a full-featured HWR model rather than a classifier, we can make the assumption that any text, predicted incorrectly from the input image, will be reported back as a spelling mistake. This assumption allows us to estimate an upper limit for the baseline performance in the misspelling detection task using a HWR-specific metric. Let us assume that the HWR model predicts correctly $P$ out of the $T$ test words and we have a misspelling detection test set consisting of $M$ incorrectly spelled examples and $C$ correctly spelled ones, where $M$ and $C$ are derived from the mistake-to-correct ratio $m:c$:

\begin{equation}
M = \frac{m}{m+c} \cdot T\label{eq:C}
\end{equation}

\begin{equation}
C = \frac{c}{m+c} \cdot T\label{eq:M}
\end{equation}

After substituting $TP=M$, $FP=C \cdot \frac{T-P}{T}$, and $FN=0$, the baseline precision $P_b$ and recall $R_b$ are expressed as:
\begin{equation}
P_b \approx \frac{m}{m + c \cdot \frac{T-P}{T}}\label{eq:P_b}
\end{equation}

\begin{equation}
R_b \approx 100\%\label{eq:R_b}
\end{equation}
As it is seen from \eqref{eq:P_b}, the zero-FN assumption has no effect on calculating the baseline precision.

\subsection{Test results}\label{subsec12}

In Table~\ref{mainResults} we compare the performance of the two methods in each assessment scenario. The precision of the baseline is calculated using \eqref{eq:P_b} given that during our experiments the HWR model predicted correctly $P=9,271$ out of the $T=15000$ handwritten words. The classifier model was calibrated to meet the minimal recall requirement of 99\%.

Our findings show that the precision of our model exceeds the baseline precision by 19.5\% on average with the improvement ranging from 14\% to 25\% depending on the difficulty of the exercise (difficult to moderate).

\begin{table}[htbp]
\caption{Performance comparison of the binary classifier against the HWR baseline in different assessment scenarios}\label{mainResults}
\begin{tabular*}{\columnwidth}{@{\extracolsep\fill}lrr}
\toprule%
                   & \multicolumn{2}{@{}c@{}}{Precision} \\
\cmidrule{2-3}
                   & Difficult scenario             & Moderate scenario\\
\midrule
HWR baseline      &                56.70\%          &       34.37\%\\
Ours              &                64.59\%          &       43.11\%\\
(improvement)     &                (+14\%)          &       (+25\%)\\
\botrule
\end{tabular*}
\end{table}

\subsection{Effect of mistake severity}\label{subsec13}

In Table~\ref{severityImpact} we demonstrate the impact that severity of mistakes has on our model’s performance. The more mistakes there are in a word, the more ``confident'' the model is, which results in increased precision. Subsequently, we can expect even higher performance in knowledge assessment exercises where an incorrect answer is typically a different word rather than a misspelling.

\begin{table}[htbp]
\caption{Classifier precision by mistake severity}\label{severityImpact}
\begin{tabular*}{\columnwidth}{@{\extracolsep\fill}lccc}
\toprule%
                    & \multicolumn{3}{@{}c@{}}{Mistake Severity}\\
\cmidrule{2-4}
                    & Level 1   & Level 2   & Level 3\\
\midrule
Difficult scenario  &     52.0\%&     64.4\%&       77.4\%\\
Moderate scenario   &     29.7\%&     45.6\%&       54.0\%\\
\botrule
\end{tabular*}
\end{table}

\subsection{Recall requirement}\label{subsec14}

\begin{figure}[htbp]
\includegraphics[width=0.48\textwidth]{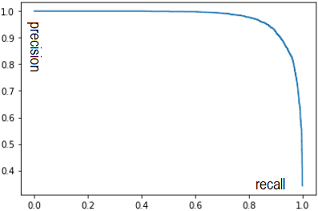}
\caption{Precision-recall curve (Difficult Assessment, Error Severity 2)}\label{difficultCurve}
\end{figure}

\begin{figure}[htbp]
\includegraphics[width=0.48\textwidth]{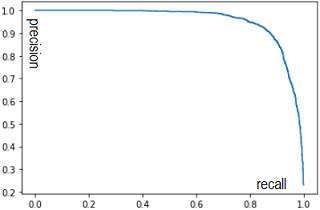}
\caption{Precision-recall curve (Moderate Assessment, Error Severity 2)}\label{moderateCurve}
\end{figure}

The precision-recall curve plots (Figure~\ref{difficultCurve} and Figure~\ref{moderateCurve}) help to gain a better understanding of the possible trade-offs when calibrating the classifier for tasks with different recall requirements.

\subsection{Future experiments}\label{subsec15}

The robustness of the evaluation could be improved by collecting and utilizing a test set with annotated handwritten misspellings (subject to consent from the participants). Other aspects to investigate include model architectures and different handwriting databases. HWR models, such as \cite{b4_2,b11,b12,b13,b17} and others, can be used as feature extractors or fine-tuned. The convolutional classification head is another area of interest where neural architectures from a variety of application domains can be explored. Since our method in principle can train on any HWR dataset or alphabet, it could be evaluated using alternative datasets such as \cite{b4_1} (images taken using mobile phone) or \cite{b38} (French).

\section{Conclusion}\label{sec5}

In this work we proposed a novel method for offline validation of handwriting against expected text. Contrary to the conventional approach, which predicts the handwritten text and compares it with the expected, we propose a binary classification neural architecture for solving the problem in a single step. Our binary classifier uses a pre-trained HWR model as a feature extractor and a convolutional classification head combining the HWR feature matrix with the one-hot matrix representing the input text. To the best of our knowledge, this type of multimodal convolution has not been documented anywhere in the HWR literature.

Due to the lack of public datasets with incorrectly spelled handwritten text, we train and test our classifier on a HWR database where the negative training examples are generated by injecting noise in the original benchmark dataset. The model's performance is evaluated in a hypothetical school assessment scenario against the conventional baseline method. The results show that our classifier can be calibrated to outperform the baseline on average by 19.5\% in multiple applications involving the validation of handwritten text. This can be particularly beneficial in the context of human-in-the-loop automation, where the workload to people can be reduced significantly depending on the problem type and difficulty.

\bibliographystyle{sn-nature}
\bibliography{refs}

\end{document}